\documentstyle[aclap,harvard]{article}

\title{\vspace{-0.5in}Representing Constraints with Automata}
\author{Frank Morawietz \and Tom Cornell\\
  Seminar f\"ur Sprachwissenschaft\\
  Universit\"at T\"ubingen\\
  Wilhelmstr.~113\\
  72074 T\"ubingen, Germany\\
  {\tt \{frank,cornell\}@sfs.nphil.uni-tuebingen.de}}

\begin{document}
\maketitle
\vspace{-0.5in}
\begin{abstract}
  In this paper we describe an approach to constraint-based syntactic
  theories in terms of finite tree automata. The solutions to
  constraints expressed in weak monadic second order (MSO) logic are
  represented by tree automata recognizing the assignments which make
  the formulas true. We show that this allows an efficient
  representation of knowledge about the content of constraints which can
  be used as a practical tool for grammatical theory verification.  We
  achieve this by using the intertranslatability of formulas of MSO
  logic and tree automata and the embedding of MSO logic into a
  constraint logic programming scheme.  The usefulness of the approach
  is discussed with examples from the realm of Principles-and-Parameters
  based parsing.
\end{abstract}

\section{Introduction}
\label{sec:intro}

In recent years there has been a continuing interest in computational
linguistics in both model theoretic syntax and finite-state techniques.
In this paper we attempt to bridge the gap between the two by exploiting
an old result in logic, that the weak monadic second order (MSO) theory
of two successor functions (WS2S) is decidable
\cite{thatcher&wright:68,doner:70}. A ``weak'' second order theory is
one in which the set variables are allowed to range only over finite
sets.  There is a more powerful result available: it has been shown
\cite{rabin:69} that the strong monadic second order theory (variables
range over infinite sets) of even countably many successor functions is
decidable.  However, in our linguistic applications we only need to
quantify over finite sets, so the weaker theory is enough, and the
techniques correspondingly simpler.\footnote{ All of these are
  generalizations to trees of results on strings and the monadic second
  order theory of one successor function originally due to
  \citeasnoun{buec:weak60}. The applications we mention here could be adapted
  to strings with finite-state automata replacing tree automata. In
  general, all the techniques which apply to tree automata are
  straightforward generalizations of techniques for FSAs.  } The
decidability proof works by showing a correspondence between formulas in
the language of WS2S and tree automata, developed in such a way that the
formula is satisfiable iff the set of trees accepted by the
corresponding automaton is nonempty. While these results were well
known, the (rather surprising) suitability of this formalism as a
constraint language for Principles-and-Parameters (P\&P) based
linguistic theories has only recently been shown by
\citeasnoun{roge:stud94}. 

It should be pointed out immediately that the translation from formulas
to automata, while effective, is just about as complex as it is possible
to be. In the worst case, the number of states can be given as a
function of the number of variables in the input formula with a stack of
exponents as tall as the number of quantifier alternations in the
formula.  However, there is a growing body of work in the computer
science literature motivated by the success of the {\sf MONA}
decision procedure \cite{klar:mona95}\footnote{ The current version of
  the {\sf MONA} tool works only on the MSO logic of strings. There
  is work in progress at the University of Aarhus to extend {\sf MONA\/}
  to ``{\sf MONA++}'', for trees \cite{bieh:guid96}.  } on the
application of these techniques in computer science
\cite{basi:hard95,kelb:mose97}, which suggests that in practical cases
the extreme explosiveness of this technique can be effectively
controlled. It is one of our goals to show that this is the case in
linguistic applications as well.

The decidability proof for WS2S is inductive on the structure
of MSO formulas. Therefore we can choose our particular tree description
language rather freely, knowing (a) that the resulting logic will be
decidable and (b) that the translation to automata will go through as
long as the atomic formulas of the language represent relations which
can be translated (by hand if necessary) to tree automata. We will see
how this is done in the next section, but the point can be appreciated
immediately. For example, \citeasnoun{nieh:feat92} and
\citeasnoun{ayar:lisa97} have investigated the
usefulness of these techniques in dealing with feature trees which
unfold feature structures; there the attributes of an attribute-value
term are translated to distinct successor functions. On the other hand,
\citeasnoun{roge:amod96} has developed a language rich in long-distance
relations (dominance and precedence) which is more appropriate for work
in Government-Binding (GB) theory.  Compact automata can be easily
constructed to represent dominance and precedence relations. One can
imagine other possibilities as well: as we will see, the automaton for
Kayne-style asymmetric, precedence-restricted c-command
\cite{kayn:anti94} is also very compact, and makes a suitable
primitive for a description language along the lines developed by
\citeasnoun{fran:ccom95}.

The paper is organized as follows. First we present some of the
mathematical background, then we discuss (na\"{\i}ve) uses of the
techniques, followed by the presentation of a constraint logic
programming-based extension of MSO logic to avoid some of the problems
of the na\"{\i}ve approach, concluding with a discussion of
its strengths and weaknesses.

\section{Defining Automata with Constraints}
\label{sec:math}

\paragraph{Tree automata.}
For completeness, we sketch the definitions of trees and tree automata
here. An introduction to tree automata can be found in
\citeasnoun{gecseg&steinby:84}, as well as in
\citeasnoun{thatcher&wright:68} and \citeasnoun{doner:70}.

Assume an alphabet $\Sigma = \Sigma_0 \cup \Sigma_2$ with $\Sigma_0 =
\{\lambda\}$ and \(\Sigma_2 \) being a set of binary operation symbols.
We think of (binary) trees over $\Sigma$ as just the set of terms
$T_\Sigma$ constructed from this alphabet.  That is, we let $\lambda$ be
the empty tree and let $\sigma(t_1,t_2)$, for $\sigma \in \Sigma_2$ and
$t_1,t_2\in T_\Sigma$, denote the tree with label $\sigma$ and subtrees
$t_1,t_2$. Alternatively, we can think of a tree $t$ as a function from
the addresses in a binary tree domain $T$ to labels in
$\Sigma$.\footnote{The first approach is developed in
  \citeasnoun{thatcher&wright:68}, the second in \citeasnoun{doner:70}. A tree
  domain is a subset of strings over a linearly ordered set which is
  closed under prefix and left sister.}

A deterministic (bottom-up) tree automaton ${\cal A}$ on binary trees is
a tuple $\langle A, \Sigma, a_0, F, \alpha \rangle$ with $A$ the set of
states, $a_0\in A$ the initial state, $F\subseteq A$ the final states
and $\alpha: (A\times A\times \Sigma) \rightarrow A$ the transition
function. The transition function can be thought of as a homomorphism
on trees
inductively defined as: $h_\alpha(\lambda) = a_0$ and
\( h_\alpha(\sigma(t_1,t_2)) = \alpha(h_\alpha(t_1),h_\alpha(t_2),\sigma) \). 
An automaton ${\cal A}$
accepts a tree $t$ iff $h_\alpha(t) \in F$. The language recognized by
 ${\cal A}$ is denoted by 
 $T({\cal A})= \{ t\, | \, h_\alpha(t) \in F\}$. 

Emptiness of the language $T({\cal A})$ is decidable by a fixpoint
construction computing the set of reachable states.
The reachability algorithm is given below in
Figure~\ref{fig:reachable}. $R$ contains the reachable states
constructed so far, and $R'$ contains possibly new states constructed
on the current pass through the loop.
\begin{figure}[htb]
\begin{tabular}{ll}
1. &\(R := \{ a_0 \}\), \(R' := \emptyset\).\\
2. &For all \( (a_i,a_j) \in R\times R \),
    for all \( \sigma \in \Sigma \), \\
    &\hspace*{2em}\( R' := R' \cup \{ \alpha(a_i,a_j,\sigma) \} \).\\
3. &If \( R' - R = \emptyset \) then return $R$,\\
    &else \( R := R \cup R' \), go to step 2.
\end{tabular}
\caption{Reachable states algorithm.}
\protect\label{fig:reachable}
\end{figure}
\( T({\cal A}) \) is empty if and only if no final state is reachable.
Naturally, if we want to test emptiness, we can stop the construction as
soon as we encounter a final state in $R'$.  Note that, given an
automaton with $k$ states, the algorithm must terminate after at most
$k$ passes through the loop, so the algorithm terminates after at most
\( k^3 \) searches through the transition table.

Sets of trees which are the language of some tree automaton are called
{\em recognizable}.\footnote{
  The recognizable sets of trees yield the context free string
  languages, so MSO logics are limited to context free power.
  However, the CLP extension discussed below can be used to 
  amplify the power of the formalism where necessary.
  }
The recognizable sets are closed under the boolean
operations of conjunction, disjunction and negation, and the automaton
constructions which witness these closure results are absolutely
straightforward generalizations of the corresponding better-known
constructions for finite state automata.
The
recognizable sets are also closed under projections (mappings from one
alphabet to another) and inverse projections, and again the construction
is essentially that for finite state automata. The projection
construction yields a nondeterministic automaton, but, again as for
FSA's, bottom-up tree automata can be made deterministic by a
straightforward generalization of the subset construction. (Note that
{\em top-down\/} tree automata do not have this property: deterministic
top-down tree automata recognize a strictly narrower family of tree
sets.) Finally, tree automata can be minimized by a construction which
is, yet again, a straightforward generalization of well known FSA
techniques.

\paragraph{The weak second order theory of two successor functions.}

One attraction of monadic second order tree logics is that
they give us a principled means of generating automata from
a constraint-based theory. The connection allows the linguist to
specify ideas about natural language in a concise manner in
logic, while at the same time providing a way of ``compiling''
those constraints into a form
which can be efficiently used in natural language processing applications.

The translation is provided via the weak monadic second order theory of
two successor functions (WS2S). The structure of two
successor functions, \( {\cal N}_2 \), has for its domain ($N_2$) the
infinite binary branching tree.  Standardly the language of WS2S is
based on two successor functions (left-daughter and right-daughter),
but, as \citeasnoun{roge:stud94} shows, this is intertranslatable with a
language based on dominance and precedence relations. Because we choose
the monadic second order language over whichever  of these two signatures
is preferred, we can quantify over sets of nodes in $N_2$.
So we can use these
sets to pick out arbitrarily large finite trees embedded in $N_2$.
Second order variables can also be used to pick out other properties of
nodes, such as category or other node-labeling features,
and they can be used to pick out higher order substructures 
such as $\bar{\textrm{X}}$ projections or chains.

As usual, satisfiability of a formula in the language of WS2S by \(
{\cal N}_2 \) is relative to an assignment function, mapping individual
variables to members of $N_2$ (as in first order logic) and mapping
monadic predicate variables to subsets of $N_2$.  Following
\citeasnoun{buec:weak60}, \citeasnoun{doner:70} and \citeasnoun{thatcher&wright:68} show
that assignment functions for such formulas can be coded by a labeling
of the nodes in $N_2$ in the following way.  First, we treat individual
variables as set variables which are constrained to be singleton sets
(we can define the singletonhood property in MSO tree logic). So,
without loss of generality, we can think of the domain of the assignment
function as a sequence \(X_1,\ldots, X_n \) of the variables occurring in
the given formula.  We choose our labeling alphabet to be the set of
length $n$ bit strings: \( \{ 0, 1 \}^n \).  Then, for every node $n \in
N_2$, if we intend to assign $n$ to the denotation of $X_i$, we indicate
this by labeling $n$ with a bit string in which the $i\/$th bit is on.
(In effect, we are labelling every node with a list of the sets to
which it belongs.)
Now every assignment function we might need corresponds uniquely to a labeling
function over $N_2$.  What Doner, and Thatcher and Wright 
(and, for strong S2S, Rabin) show is that each formula in the language of WS2S
corresponds to a tree automaton which recognizes just the satisfying
``assignment labelings'', and we can thereby define a notion of
``recognizable relation''.  So the formula is satisfiable just in case
the corresponding automaton recognizes a nonempty language.  Note that
any language whose formulas can be converted to automata in this way is
therefore guaranteed to be decidable, though whether it is as strong
as the language of
WS2S must still be shown. 

This approach to theorem-proving is rather different from more general
techniques for higher-order theorem proving in ways that the
formalizer must keep in mind. In particular, we are deciding
membership in the theory of a fixed structure, ${\cal N}_2$, and
not consequence of an explicit set of tree axioms. So, for example,
the parse tree shows up in the formalization as a second order
variable, rather than simply being a satisfying model
(cf.~\citeasnoun{johnson:l&p}, on ``satisfiability-based''
grammar formalisms).

As an example consider the following formula denoting the relation of
directed asymmetric c-command\footnote{
  This relation is not monadic, but
  reducible via syntactic substitution to an MSO signature. In fact, we
  can define relations of any arity as long as they are explicitly
  presentable in MSO logic.
  } 
in the sense of \citeasnoun{kayn:anti94}.  We use
the tree logic signature of \citeasnoun{roge:stud94}, which, in a second order
setting, is interdefinable with the language of multiple successor
functions.  Uppercase letters denote second order variables, lowercase
ones first order variables, $\lhd^{*}$ reflexive domination, $\lhd^{+}$
proper domination and $\prec$ proper precedence:
\begin{displaymath}
  \begin{array}{l}
    \mbox{{\it AC-Com\/}}(x,y) \stackrel{def}{\Longleftrightarrow}
    \\
    \phantom{\mbox{{\it AC}}}\hspace*{2em}\mbox{\% $x$ c-commands
    $y$:}\\ 
    \phantom{\mbox{{\it AC}}}(\forall z)[z\lhd^+x \Rightarrow
    z\lhd^+y] \wedge\neg (x\lhd^* y) \wedge\null  
    \\
    \phantom{\mbox{{\it AC}}}\hspace*{2em}\mbox{\% $y$ does not
    c-command $x$:}\\ 
    \phantom{\mbox{{\it AC}}}\neg((\forall z)[z\lhd^+y \Rightarrow
    z\lhd^+x] \wedge\neg (y\lhd^* x)) 
    \wedge\null\\
    \phantom{\mbox{{\it AC}}}\hspace*{2em}\mbox{\% $x$ preceeds
    $y$:}\\ 
    \phantom{\mbox{{\it AC}}}x\prec y
  \end{array}
\end{displaymath}

The corresponding tree automaton is shown in Figure \ref{aut:ac-comm}.
On closer examination of the transitions, we note that
\begin{figure}[htb]
  \begin{center}
    \leavevmode
    \( {\cal A} = \langle A, \Sigma, a_0, F, \alpha \rangle, \)\\
    \( A = \{ a_0, a_1, a_2, a_3, a_4, a_5 \}, \) \\ 
    \( \Sigma = \{ \underline{1}\underline{1}, \underline{1}\underline{0},
    \underline{0}\underline{1}, \underline{0}\underline{0} \} \) \\
    \( F = \{ a_4 \}, \)\\[2ex]
    $\alpha(a_0,a_0,\underline{0}\underline{0}) = a_0$ \hspace*{.5cm}
    $\alpha(a_0,a_0,\underline{1}\underline{0}) = a_3$\\
    $\alpha(a_0,a_0,\underline{0}\underline{1}) = a_1$ \hspace*{.5cm}
    $\alpha(a_0,a_1,\underline{0}\underline{0}) = a_2$\\
    $\alpha(a_0,a_2,\underline{0}\underline{0}) = a_2$ \hspace*{.5cm}
    $\alpha(a_0,a_4,\underline{0}\underline{0}) = a_4$\\
    $\alpha(a_1,a_0,\underline{0}\underline{0}) = a_2$ \hspace*{.5cm}
    $\alpha(a_2,a_0,\underline{0}\underline{0}) = a_2$\\
    $\alpha(a_3,a_2,\underline{0}\underline{0}) = a_4$ \hspace*{.5cm}
    $\alpha(a_4,a_0,\underline{0}\underline{0}) = a_4$\\
    all other transitions are to $a_5$
    \caption{The automaton for $AC\mbox{-}Com(x,y)$}
    \label{aut:ac-comm}
  \end{center}
\end{figure}
we just percolate the initial state as long as we find only nodes which
are neither $x$ nor $y$. From the initial state on both the left and the
right subtree we can either go to the state denoting ``found $x$''
($a_3$) if we read symbol $\underline{1}\underline{0}$ or to the state
denoting ``found $y$'' ($a_1$) if we read symbol
$\underline{0}\underline{1}$. After finding a dominating node -- which
switches the state to $a_2$ -- we can then percolate $a_2$ as long as
the other branch does not immediately dominate $x$.  When we have $a_3$
on the left subtree and $a_2$ on the right one, we go to the final state
$a_4$ which again can be percolated as long as empty symbols are read.
Clearly, the automaton recognizes all trees which have the desired
c-command relation between the two nodes.  It compactly represents the
(infinite) number of possible satisfying assignments.  The proof of the
decidability of WS2S furnishes a technique for deriving such automata
for recognizable relations effectively. (In fact the above automaton was
constructed by a simple implementation of such a compiler which we have
running at the University of T\"ubingen. See \citeasnoun{mora:recog96}.)
The proof is inductive.  In the base case, relations defined by atomic
formulas are shown to be recognizable by brute force. Then the induction
is based on the closure properties of the recognizable sets, so that
logical operators correspond to automaton constructions in the following
way: conjunction and negation just use the obvious corresponding
automaton operations and existential quantification is implemented with
the projection construction.  The inductive nature of the proof allows
us a fairly free choice of signature, as long as our atomic relations
are recognizable. We could, for example, investigate theories in which
asymmetric c-command was the only primitive, or asymmetric c-command
plus dominance, for example.

The projection construction, as noted above, yields
nondeterministic automata as output, and the negation construction
requires deterministic automata as input, so the subset construction
must be used every time a negated existential quantifier is
encountered. The corresponding exponential blowup in the state space
is the main cause of the non-elementary complexity of the
construction. Since a quantifier prefix of the form
\( \exists\cdots\exists\forall\cdots\forall\exists\cdots \)
is equivalent to 
\( \exists\cdots\exists\neg\exists\cdots\exists\neg\exists\cdots \)
we see that the stack of exponents involved is determined by the
number of quantifier alternations.

It is obviously desirable to keep the automata as small as possible.
In our own prototype, we minimize the outputs of all of our
automata constructions. 
Note that this gives us another way of determining satisfiability,
since the minimal automaton recognizing the empty language is readily
detectable: its only state is the initial state, and it is not final.

\section{Defining Constraints with Automata}
\label{sec:offline}

An obvious goal for the use of the discussed approach would be the
(off\/line) generation of a tree automaton representing an entire
grammar.  That is, in principle, if we can formalize a grammar in an MSO
tree logic, we can apply these compilation techniques to construct
an automaton which recognizes all and only the valid parse trees.\footnote{
  This is reminiscent of approaches associated with Bernard Lang.
  See \citeasnoun{noor:thei95} and references therein.
  }
In
this setting, the parsing problem becomes the problem of conjoining an
automaton recognizing the input with the grammar automaton, with the
result being an automaton which recognizes all and only the valid parse
trees.
For example, assume that we have an automaton $Gram(X)$ such that $X$
is a well-formed tree, and suppose we want to recognize the input
{\em John sees Mary}. Then we conjoin a description of the input with
the grammar automaton as given below.
\begin{displaymath}
  \begin{array}{l}
    (\exists x,y,z \in X) [x\in John \wedge y\in Sees \wedge z\in Mary
    \wedge \null\\
    \hspace*{.5cm} x\prec y\prec z \wedge Gram(X)]
  \end{array}
\end{displaymath}
The recognition problem is just the problem of determining whether or
not the resulting automaton recognizes a nonempty language.  Since the
automaton represents the parse forest, we can run it to generate parse
trees for this particular input.  

Unfortunately, as we have already
noted, the problem of generating a tree automaton from an arbitrary MSO
formula is of non-elementary complexity.  Therefore, it seems unlikely
that a formalization of a realistic principle-based grammar could be
compiled into a tree automaton before the heat death of the universe.
(The formalization of ideas from {\em Relativized Minimality}
\cite{rizz:rela90} presented in \citeasnoun{roge:stud94} fills an entire
chapter without specifying even the beginning of a full lexicon, for
example.)
Nonetheless there are a number of ways in which these compilation
techniques remain useful. 
First, though the construction of a grammar automaton is almost
certainly infeasible for realistic grammars, the construction of a
grammar-and-input automaton---which is a very much smaller
machine---may not be. We discuss techniques based on constraint logic
programming that are  applicable to that problem in the next section. 
 
Another use for such a compiler is
suggested by the standard divide-and-conquer strategy for problem
solving: instead of compiling an entire grammar formula, we isolate
interesting subformulas, and attempt to compile them. 
Tree automata represent properties of trees and there are many such
properties less complex than global well-formedness which are
nonetheless important to establish for parse trees. In particular,
where the definition of a property of parse trees involves negation or
quantification, including quantification over sets of nodes, it may be
easier to express this in an MSO tree logic, compile the resulting
formula, and use the resulting automaton as a filter on parse trees
originally generated by other means (e.g., by a covering phrase
structure grammar).

At the moment, at least, the question of which grammatical properties
can be compiled in a reasonable time 
is largely  empirical. It is made even more difficult by
the lack of high quality software tools. This situation should be
alleviated in the near future when work on {\sf MONA++} at the
University of Aarhus is completed; the usefulness of its older sister
{\sf MONA} \cite{klar:mona95}, 
which works on strings and FSA's, has been well
demonstrated in the computer science literature. 
In the meantime, for tests, we are using a comparatively simple
implementation of our own. Even with very low-power tools, however, we
can construct automata for interesting grammatical constraints. 

For example, recall the definition of asymmetric c-command and its
associated automaton in Figure~\ref{aut:ac-comm}. In linguistic
applications, we generally use versions of c-command which are
restricted to be local, in the sense that no element of a certain type
is allowed to intervene. The general form of such a locality condition $LC$
might then be formalized as follows.
\begin{displaymath}
  \begin{array}{l}
    \mbox{{\it LC\/}}(x,y) \stackrel{def}{\Longleftrightarrow}\\
    \phantom{\mbox{{\it LC}}}\mbox{{\it AC-Com}}(x,y) \wedge\null \\
    \phantom{\mbox{{\it LC}}}\hspace*{1.5em}\mbox{\% there does not
    exist $z$ with property $P$:}\\ 
    \phantom{\mbox{{\it LC}}}(\neg\exists z)[z\in P \wedge\null \\
    \phantom{\mbox{{\it LC}}}\hspace*{1.5em}\mbox{\% such that it
    intervenes between $x$ and $y$:}\\ 
    \phantom{\mbox{{\it LC}}}\hspace{1em}(\exists w)[w\lhd x \wedge
    w\lhd^+z \wedge z\lhd^+y]] 
  \end{array}
\end{displaymath}
Here property $P$ is meant to be the property identifying a relevant
intervener for the relation meant to hold between $x$ and $y$. Note
that this property could include that some other node be the left
successor of $z$ with certain properties, that is, this general scheme
fits cases where the intervening item is not itself directly on the
path between $x$ and $y$. This formula was compiled by us and yields
the automaton in Figure~\ref{aut:lc-cmd}. Here the first bit position
indicates membership in $P$, the second is for $x$ and the third for
 $y$.
\begin{figure}[htb]
  \begin{center}\leavevmode
    \( {\cal A} = \langle A, \Sigma, a_0, F, \alpha \rangle, \)\\
    \( A = \{ a_0, a_1, a_2, a_3, a_4, a_5 \}, \) \\ 
    \( F = \{ a_4 \}, \)\\[2ex]
    \( \alpha(a_0 , a_0, \underline{000}) = a_0 \hspace*{0.5cm}
       \alpha(a_0 , a_0, \underline{100}) = a_0 \) \\
    \( \alpha(a_0 , a_0, \underline{010}) = a_3 \hspace*{0.5cm}
       \alpha(a_0 , a_0, \underline{110}) = a_3 \) \\
    \( \alpha(a_0 , a_0, \underline{001}) = a_1 \hspace*{0.5cm}
       \alpha(a_0 , a_0 ,\underline{101}) = a_1 \) \\
    \( \alpha(a_0 , a_1 ,\underline{000}) = a_2 \hspace*{0.5cm}
       \alpha(a_0 , a_2 ,\underline{000}) = a_2 \) \\
    \( \alpha(a_0 , a_4 ,\underline{000}) = a_4 \hspace*{0.5cm}
       \alpha(a_0 , a_4 ,\underline{100}) = a_4 \) \\
    \( \alpha(a_1 , a_0 ,\underline{000}) = a_2 \hspace*{0.5cm}
       \alpha(a_2 , a_0 ,\underline{000}) = a_2 \) \\
    \( \alpha(a_3 , a_2 ,\underline{000}) = a_4  \hspace*{0.5cm}
       \alpha(a_3 , a_2 ,\underline{100}) = a_4 \) \\
    \( \alpha(a_4 , a_0 ,\underline{000}) = a_4 \hspace*{0.5cm}
       \alpha(a_4 , a_0 ,\underline{100}) = a_4 \) \\
    all other transitions are to $a_5$
  \end{center}
  \caption{Automaton for local c-command.}
  \protect\label{aut:lc-cmd}
\end{figure}

This automaton could in turn be implemented itself as Prolog code, and
considered to be an optimized implementation of the given specification.
Note in particular the role of the compiler as an optimizer.  It outputs
a minimized automaton, and the minimal automaton is a unique (up to
isomorphism) definition of the given relation.  Consider again the
definition of AC-Command in the previous section.  It is far from the
most compact and elegant formula defining that relation.  There exist
much smaller formulas equivalent to that definition, and indeed some are
suggested by the very structure of the automaton. That formula was
chosen because it is an extremely straightforward formalization of the
prose definition of the relation.  Nonetheless, the automaton compiled
from a much cleverer formalization would still be essentially the same.
So no particular degree of cleverness is assumed on the part of the
formalizer; optimization is done by the compiler.\footnote{ The
  structure of the formula {\em does} often have an effect on the time
  required by the compiler; in that sense writing MSO formalizations is
  still Logic Programming.}

\section{MSO Logic and Constraint Logic Programming}
\label{sec:clp}

The automaton for a grammar formula is presumably quite a lot larger
than the parse-forest automaton, that is, the automaton for the grammar
conjoined with the input description.  So it makes sense to search for
ways to construct the parse-forest automaton which do not require the
prior construction of an entire grammar automaton.  In this section we
consider how we might do this by by the embedding of the MSO constraint
language into a constraint logic programming scheme.  The constraint
base is an automaton which represents the incremental accumulation of
knowledge about the possible valuations of variables.  As discussed
before, automata are a way to represent even infinite numbers of
valuations with finite means, while still allowing for the efficient
extraction of individual valuations.  We incrementally add information
to this constraint base by applying and solving clauses with their
associated constraints.  That is, we actually use the compiler {\em on
  line} as the constraint solver.  Some obvious advantages include that
we can still use our succinct and flexible constraint language, but gain
(a) a more expressive language, since we now can include inductive
definitions of relations, and (b) a way of guiding the compilation
process by the specification of appropriate programs.

In \citeasnoun{mora:ltwo96}, we define a relational extension ${\cal
  R}$(WS2S) of our constraint language following the H\"ohfeld and
Smolka scheme \cite{hoeh:defi88}. From the scheme we get a sound and
complete, but now only semi-decidable, operational interpretation of a
definite clause-based derivation process.  The resulting structure is an
extension of the underlying constraint structure with the new relations
defined via fixpoints.

As usual, a definite clause is an implication with an atom as the head
and a body consisting of a satisfiable MSO constraint and a (possibly
empty) conjunction of atoms.  A derivation step consists of  two
parts: goal reduction, which substitutes the body of a goal for an
appropriate head, and constraint solving, which means in our case that
we have to check the satisfiability of the constraint associated with
the clause in conjunction with the current constraint store. For
simplicity we assume a standard left-to-right, depth-first interpreter
for the execution of the programs. 
The solution to a search branch of a program is a satisfiable
constraint,
represented in ``solved form'' as an automaton.
Note that automata do make appropriate solved forms for systems of
constraints: minimized automata are normal forms, and they allow for
the direct and efficient recovery of particular solutions.

Intuitively, we have a language which has an operational interpretation
similar to Prolog with the differences that we interpret it not on the
Herbrand universe but on $N_2$, that we use MSO constraint
solving instead of unification and that we can use defined (linguistic)
primitives directly. 

The resulting system is only semi-decidable, due to the fact that the
extension permits monadic second order variables to appear in
recursively defined clauses.  So if we view the inductively defined
relations as part of an augmented signature, this signature contains
relations on sets.  These allow the specification of undecidable
relations; for example, \citeasnoun{mora:ltwo96} shows how to encode the PCP.
If we limit ourselves to just singleton variables in any directly or
indirectly recursive clause, every relation we define stays within the
capacity of MSO logic,\footnote{ 
  Relations on individuals describe sets
  which are expressible as monadic predicates.
  } 
since, if they are first
order inductively definable, they are explicitly second order definable
\cite{roge:stud94}.  Since this does not take us beyond the power of
MSO logic and natural language is known not to be context-free, the
extra power of ${\cal R}$(WS2S) offers a way to get past the
context-free boundary.

To demonstrate how we now split the work between the compiler and the
CLP interpreter, we present a simple example.  Consider the
following na\"{\i}ve specification of a lexicon:\footnote{Here and in the
  following we treat free variables as being stored in a global table so
  that we do not have to present them in each and every constraint. In
  particular, without this $lexicon$ would have the additional arguments
  $Sees$, $V$, $John$, $N$, $Mary$ and all free variables appearing in
  the other definitions.}
\begin{displaymath}
  \begin{array}{rcl}
    Lexicon(x) & \stackrel{def}{\Longleftrightarrow} & 
    (x\in Sees \wedge x\in V \wedge \dots)\\
    &    \vee & (x\in John \wedge x\in N \wedge \dots)\\
    &    \vee & (x\in Mary \wedge x\in N \wedge \dots)\\
    & \vdots  &
  \end{array}
\end{displaymath}
We have specified a set called $Lexicon$ via a disjunctive specification
of lexical labels, e.g.\ $Sees$, and the appropriate combination of
features, e.g.\ $V$.  Na\"{\i}vely, at least, every feature we use must
have its own bit position, since in the logic we treat features as set
variables. So, the alphabet size with the encoding as bitstrings will be
at least $2^{|Alphabet|}$. It is immediately clear that the compilation
of such an automaton is extremely unattractive, if at all feasible.

We can avoid having to compile the whole lexicon by having separate
clauses for each lexical entry in the CLP extension.  Notational
conventions will be that constraints associated with clauses are written
in curly brackets and subgoals in the body are separated by \&'s. Note
that relations defined in ${\cal R}$(WS2S) are written lowercase.
\begin{displaymath}
\begin{array}{rcl}
  lexicon(x) & \longleftarrow & \{ x\in Sees \wedge x\in V \wedge \dots \}\\
  lexicon(x) & \longleftarrow & \{ x\in John \wedge x\in N \wedge \dots \}\\
  lexicon(x) & \longleftarrow & \{ x\in Mary \wedge x\in N \wedge \dots \}\\
  & \vdots &
\end{array}
\end{displaymath}
This shifts the burden of handling disjunctions
to the interpreter.
The intuitive point should be clear: it
is not the case that every constraint in the grammar has to be
expressed in one single tree automaton. We need only compile into the
constraint store those which are really needed. Note that this is true even
for variables appearing in the global table. In the CLP extension the
appearance in the table is not coupled to the appearance in the
constraint store. Only those are present in both which are part of the
constraint in an applied clause.

We can also use off\/line compiled modules in a ${\cal R}$(WS2S) 
parsing program.  
As a source of simple examples, we 
draw on the definitions from the lectures on P\&P parsing presented in
\citeasnoun{john:cons95}.
In implementing a program such as 
Johnson's simplified $parse$ relation---see Figure \ref{defjoparse}---we
can in principle define any of the subgoals in the body either via
precompiled automata (so they are essentially treated as facts), or
else providing them with more standard definite clause definitions. 

\begin{figure}[htb]
\begin{center}
  $\begin{array}[t]{lll}
             parse(Words,Parse)& \longleftarrow      &    \\
             & \{ Tree(Parse)\} & \& \\
             & yield(Words,Parse) & \&\\
             & xbar(Parse) & \&\\
             & ecp(Parse) &\\
           \end{array}
           $
\end{center}
    \caption{$parse$ as in \citeasnoun{john:cons95}}
    \label{defjoparse}
\end{figure}

In more detail, {\it Words} denotes a set of nodes labeled according to the
input description. Our initial constraint base, which can be
automatically generated from a Prolog list of input words, is the
corresponding tree automaton.  The associated constraint $Tree$ is
easily compilable and serves as the initialization for our parse
tree.  The $yield$ and $ecp$ predicates can easily be explicitly defined
and, if practically compilable (which is certainly the case for
 $yield$), could then be treated as facts. 
The $xbar$ predicate, on the other
hand, is a disjunctive specification of licensing conditions depending
on different features and configurations, e.g., whether we are faced
with a binary-, unary- or non-branching structure, which is better
expressed as several separate rules. In fact, since we want the lexicon
to be represented as several definite clauses, we cannot have $xbar$ as
a simple constraint. This is due to the limitation of the constraints
which appear in the definite clauses to (pure) MSO constraints. 

We now have another well-defined way of using the offline compiled
modules. This, at least, separates the actual processing issues (e.g.,
$parse$) from the linguistically motivated modules (e.g., $ecp$).  One
can now see that with the relational extension, we can not only use
those modules which are compilable directly, but also guide the
compilation procedure.
In effect this means interleaving the intersection of the grammar and
the input description such that only the minimal amount of information
to determine the parse is incrementally stored in the constraint base.

Furthermore, the language of  ${\cal R}$(WS2S) is sufficiently
close to standard Prolog-like programming languages to allow the
transfer of techniques and approaches developed in the realm of
P\&P-based parsing. In other words, it needs only little effort to
translate a Prolog program to a ${\cal R}$(WS2S) one.

\section{Conclusions and Outlook}
\label{sec:conc}

In this paper we presented a first step towards the realization
of a system using automata-based theorem-proving techniques to implement
linguistic processing and theory verification.  
Despite the staggering complexity bound the
success of and the continuing work on these techniques in computer
science promises a useable tool to test formalization of grammars. The
advantages are readily apparent. The direct use of a succinct and
flexible description language together with an environment to test the
formalizations with the resulting finite, deterministic tree automata
offers a way of combining the needs of both formalization and
processing. And furthermore, the CLP extension offers an even more
powerful language which allows a clear separation of processing and
specification issues while retaining the power and flexibility of the
original. Since it allows the control of the generation process,
the addition of information to the constraint base is dependent on the
input which keeps the number of variables smaller and by this the
automata more compact. 

Nevertheless it remains to be seen how far the system can be advanced
with the use of an optimized theorem-prover. The number of variables our
current prototype can handle lies between eight and eleven.\footnote{
  Note that this corresponds to 256 to 2048 different bitstrings.  }
This is not enough to compile or test all interesting aspects of a
formalization. So further work will definitly involve the optimization
of the prototype implementation, while we await the development of more
sophisticated tools like {\sf MONA++}.  It seems to be promising to
improve the (very basic) CLP interpreter, too. The H\"ohfeld and Smolka
scheme allows the inclusion of existential quantification into the
relational extension. We intend to use this to provide the theoretical
background of the implementation of a garbage collection procedure which
projects variables from the constraint store which are either local to a
definite clause or explicitly marked for projection in the program so
that the constraint store can be kept as small as possible.

\section{Acknowledgements}
This work has been supported by the project A8 of the SFB 340 of the
Deutsche Forschungsgemeinschaft. We wish especially to thank Uwe
M\"onnich and Jim Rogers for discussions and advice. Needless to say,
any errors and infelicities which remain are ours alone. Appears in the
Proceedings of ACL/EACL '97, Madrid, Spain.

\bibliographystyle{dcu}

\end{document}